# Attempto Controlled English (ACE)

Norbert E. Fuchs, Rolf Schwitter

Department of Computer Science, University of Zurich

CH-8057 Zurich, Switzerland

{fuchs, schwitter}@ifi.unizh.ch

Attempto Controlled English (ACE) allows domain specialists to interactively formulate requirements specifications in domain concepts. ACE can be accurately and efficiently processed by a computer, but is expressive enough to allow natural usage. The Attempto system translates specification texts in ACE into discourse representation structures and optionally into Prolog. Translated specification texts are incrementally added to a knowledge base. This knowledge base can be queried in ACE for verification, and it can be executed for simulation, prototyping and validation of the specification.

## 1   Motivation

*Somewhere between ridiculous pedantry and erroneous formulation there presumably exists a reasonably precise way of specifying a problem in English [Dodd 90].*

Creating reliable software is hard. One of the worst obstacles to build a good software product grows out of shortcomings in writing a complete, consistent and unambiguous requirements specification. Managers and domain specialists often find it extraordinarily difficult to formulate specifications since at the beginning of the requirements engineering process the knowledge is usually informal, incomplete and opaque, and many – possibly conflicting – personal views of the system exists. Nobody knows what exactly the program should do until there exists a first version to run.

Requirements specifications are mostly written in natural language because they need to be understood by all participants. This involves a risk since the expressive power of unrestricted natural language can tempt people to write ambiguous or even incomprehensible statements. Apart from natural language people use arbitrary graphics, or semi-formal representations like structured analysis or entity-relationship diagrams that often have no formal semantics, or a poorly defined one, thus making formal reasoning impossible [Pohl 93].

Even when the software development team gets an acceptable requirements specification there can be problems because different people may understand the same document differently. To avoid disparate interpretations of a document, people have suggested to use formal methods [Hall 90]. However, formal languages are not easily understood by untrained users. Moreover, it is far from trivial to derive a formal specification from informal requirements since this derivation process cannot be formalised and cannot be formally validated [Hoare 87]. In the end, natural language comes back in through the back door when the formal specification must be accompanied by a natural language description that paraphrases *'what the specification means in real-world terms and why the specification says what is does'* [Hall 90]. It seems that introducing formal methods into the predominantly creative process of software development runs into immense difficulties.

But there is a way out. The specification language Attempto Controlled English (ACE) combines the familiarity of natural language with the rigor of formal languages. ACE enforces writing standards that restrict the grammar and the vocabulary, thus leading to documents containing more predictable and less ambiguous language. ACE helps people to find an agreement about the correct interpretation of a requirement specification. When domain specialists and software developers are guided to use the same word for the same concept in a consistent way then misunderstandings can be reduced. ACE can be accurately and efficiently processed by a computer and is designed so that a text can be represented unambiguously in first-order predicate logic. The translated document can be verified for completeness and consistency by querying it in ACE, and by executing it. Thus validation and prototyping in concepts close to the application domain become possible and the results can be understood by all parties concerned.

Controlled or simplified English is not a new idea. It has been used for quite some time for technical documentation [Wojcik et al. 90, Adriaens & Schreurs 92, AECMA 95], and recently both for knowledge-based machine translation in the KANT system [Mitamura & Nyberg 95] and as data base query language [Androutsopoulos 95]. Additionally, a general computer processable controlled language has been suggested that could be used for various purposes ranging from structured documentation over access to information to the control of devices [Pulman & Rayner 94]. However, very few researchers have tried to employ controlled natural language for requirements specifications since this leads to further syntactic and semantic constraints for the language, especially if one requires the specifications to be executable [Ishihara et al. 92, Macias & Pulman 92, Pulman 94, Fuchs & Schwitter 95].

Habitability of a controlled subset of natural language seems to be achievable, particularly when the system gives feedback to domain specialists of the analysed sentences in a paraphrased form using the same controlled language [Capindale & Crawford 89]. Of course, before domain specialists can use controlled languages, they must be trained – in this respect ACE is not different from other methods.

## 2     Characteristics of ACE

ACE is a computer processable subset of English for writing requirements specifications. Using ACE does not presuppose expertise in formal methods or computational linguistics. With ACE we offer domain specialists an application-specific language that breaks the bottleneck between informal and formal specification methods. By making true statements about the domain of discourse, domain specialists can express their concepts in ACE in a direct and natural way using the objects of the language as abstract entities. Specifications written in ACE are textual views of formal specifications in logic. They give the impression of being informal though the language is in fact formal and machine executable.

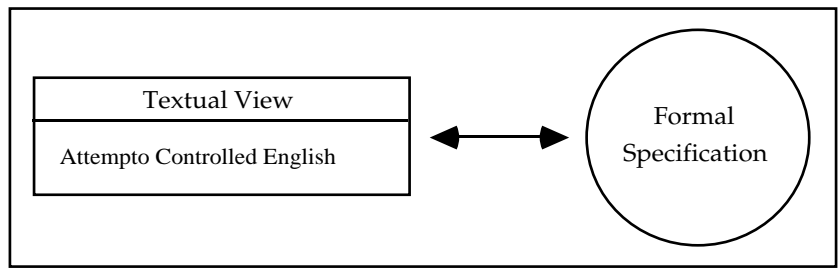

ACE is sufficiently expressive to write specifications of high quality and high readability and understandability. This is essential because the development of a specification is a cyclic activity. The specification will be written, read, interpreted, criticised, and rewritten, again and again until a satisfactory result is produced [Sommerville 92].

The language ACE is embedded in the specification system Attempto that accepts specification texts and translates them into discourse representation structures, and optionally into Prolog. Parsing in Attempto is deterministic and resembles parsing a programming language. The parser of the current Attempto system generates a syntax tree as syntactic representation, and concurrently a discourse representation structure (DRS) as semantic representation. A DRS is a structured form of first-order predicate logic which contains discourse referents representing the objects of the discourse, and conditions for these discourse referents. Each sentence is translated in the context of the preceding sentences, yielding an extension of the current DRS into which the present interpretation step is incorporated. At the present time there is no support for general reasoning in Attempto's knowledge base to resolve any ambiguity. It may, however, be necessary to add such a support as the language evolves, e.g. if a type hierarchy is introduced.

To inform the user about the results of the analysis, the parser generates a paraphrase in ACE – displaying all substitutions and interpretations made – that explains how Attempto interpreted the input text. It is up to the user to accept the interpretation, or to rephrase the input to achieve another interpretation. Additionally the user can query the knowledge base in ACE. Questions are translated into query DRSs and answered by deduction.

ACE provides a set of principles and recommendations to constrain the vocabulary and the grammar for a specification text. There are two goals behind this strategy. The first goal is to reduce the amount of lexical, structural and semantic ambiguity prior to language analysis so that the meaning of the specification text can finally be represented unambiguously. The second goal is to encourage a clear and visual writing style for the communication between the domain specialist and the software developer.

## 3  The Language ACE

The above design decisions have led to specific constraints for the language ACE where general linguistic theories have little direct help to offer – at best some broad guidelines.

### 3.1  Vocabulary

Lexical coverage is critical for robust text analysis. Though the vocabulary of ACE contains entries of the function word class, e.g. determiners, prepositions, pronouns and conjunctions, the entries for domain specific subsets of the content word class, e.g. nouns, verbs, adjectives and adverbs have to be added as needed for the specification text. Unlike function words, members of the content word class change over time as words are borrowed or new words and phrases are coined to express new technical concepts. Certain adverbs that are used for structural disambiguation, e.g. *each* and *together*, are already predefined.

To take these facts into account, Attempto's lexical editor allows domain specialists to incrementally and interactively modify and extend the lexicon for content words while the system parses the specification text. The expert interface to the lexicon represents lexical entries as complete feature structures and allows experts to freely modify any lexical entry. The interface for non-experts employs annotated templates that help users with minimal linguistic knowledge to enter information. If, for example, a non-expert adds a noun to the lexicon information about number, gender and type – countable noun or mass noun – is sufficient. Besides, the user is free to associate one or more synonyms or abbreviations with the entry.

Attempto shifts the responsibility to check for lexical ambiguity onto the users; they decide how to use a word in a specification text – and bear the consequences.

*3.1.1 Nouns*

In ACE we distinguish three subclasses of nouns: common nouns, proper nouns and personal pronouns. Common nouns are one-word nouns or compounds formed from a sequence of words. Attempto does not attempt to derive the meaning of a compound noun from the meanings of its components. Therefore, the user must enter compounds into the lexicon where they are completely lexicalised.

All common nouns fall into one of two subclasses: they occur as countable nouns or as mass nouns. If a noun is countable it can be used with an indefinite article or with numbers in front of it. Thus countable nouns distinguish singular from plural forms and can be used in questions like *How many?*. Mass nouns use neither indefinite articles nor numbers. Moreover, they do not have plural forms. It is important to notice that the majority of nouns can be used with either kind of interpretation.

*3.1.2 Verbs*

Verbs are the central words in ACE denoting states and events. Verbs can only be used actively, in the simple present tense, and in their third person singular and plural forms. In this way, users can only express statements that are always true and refer to events or states which are true in the present period of time, or possibly true when used in conditional sentences. The active form of verbs also forces users to make causal agents explicit.

Furthermore, with this restriction we achieve that the domain specialist organises the specification as step-by-step instructions – establishing the correct chronological order of events – and we exclude thereby needless complexity of temporal references. This is a gain, if we remember that tense in English is often only loosely related to time [Kamp & Reyle 93].

Some English verbs are constructed with the help of a particle, e.g. *turn on*. Particles usually overlap with prepositions leading to ambiguous readings for the same sentence. A strategy to remedy this ambiguous reading is to choose a one-word verb, e.g. *start* instead of *turn on* [Mitamura & Nyberg 95].

Modal verbs are banished from ACE because they can be used to express a vague degree of certainty about facts. The scale of vagueness extends from greatest uncertainty (*might*) to the greatest certainty (*must*). Modal verbs in a specification text are an indicator that the writer does not have a clear understanding of the state of affairs.

*3.1.3 Adjectives*

Adjectives have the function of complements (predicative use) or modifiers in noun phrase (attributive use). Semantically, adjectives denote properties. In ACE the only case of degree modification allowed is comparison, and this can be expressed either inflectionally, or analytically by means of the degree adverb *more* and *most*. Regular comparative and superlative forms are generated automatically by the lexical editor while irregular forms must be entered by the user.

The only participal forms that are accepted in ACE are de-participal adjectives in attributive and predicative position.

Modal adjectives have no business in ACE specifications. Adjectives like *possible*, *probable*, *certain*, *sure* and *necessary* modify a state of affair across a large dimension of modalities rather than adding some concrete information.

*3.1.4 Numbers and Technical Symbols*

Numbers with or without units must be entered in a fixed format. Natural numbers up to ten can be spelled out.

## 3.2 Grammar

*3.2.1 ACE Specifications*

The basic construct of an ACE specification is the declarative sentence. Strictly speaking our use of the term *declarative sentence* comes from speech act theory and has two aspects: a propositional content and an illocutionary force. A declarative sentence tells us how the world looks like if the sentence is true (proposition) and claims that the world looks like that (illocution). This is exactly what is meant by a specification.

Declarative sentences can be combined by constructors to powerful composite sentences, while certain forms of anaphora and ellipsis leave the language concise and natural. Furthermore, we place interrogative sentences at the user's disposal for verifying the translated specification text.

We suggest the following basic model of ACE as a language for requirements specifications that is expressive enough to allow natural usage and can be accurately and efficiently processed by a computer.

Specification texts consist of

- declarative sentences *subject + finite verb (+ complement or object)*
- composite sentences built from simpler sentences with the help of constructors that mark coordination (*and*, *or*, *either-or*), subordination (*if-then*, *who/which/that*), negation (*not*), and negated coordination (*neither-nor*)

Sentences can contain

- subject and object modifying relative sentences
- anaphoric references, e.g. personal pronouns
- coordination between equal constituents, e.g. *and, or*
- ellipsis as reduction of coordination
- negated noun phrases, *no X*
- synonyms and abbreviations

Interrogative sentences comprise

- *yes/no* questions
- *wh*-questions

Similar constructs have been proposed for the *computer-processable natural language* of Pulman and collaborators [Macias & Pulman 92, Pulman 94].

Here is a small excerpt of the ACE specification of a simple automated teller machine called SimpleMat.

```
The customer enters a card and a numeric personal code.
If it is not valid then SM rejects the card.
```

The example specification text employs

- composite sentences built from declarative sentences with the help of the constructors *and*, *if-then* and *not*
- ellipsis
- compound nouns, e.g. *personal code*
- anaphoric references via the pronoun *it* and the definite noun phrase *the card*
- abbreviations (*SM* standing for the name SimpleMat)

*3.2.2 Anaphora and their resolution*

In ACE anaphora resolution is a process of syntactic reconstruction and restricted to noun phrase anaphora explicitly mentioned in the previous text. In the simplest case, the anaphor is a personal pronoun and its antecedent is a noun phrase that precedes it. In order to resolve a pronominal anaphoric reference, Attempto looks back in the search space to find the most recent antecedent that agrees in gender and number. Thus, the user must use personal pronouns in a principled way referring always to the last appropriate antecedent.

Another case of anaphora are definite noun phrases that refer to discourse referents introduced existentially in the previous context. Here too, Attempto's algorithm for reference resolution searches backward until the first suitable antecedent is found. If no antecedent can be found for a definite noun phrase then – following Russell's classic analysis – a unique reference use is assumed.

### 3.2.3 *Ellipsis*

Similar to the resolution of anaphora, ellipsis – used to reduce coordination – is handled by syntactical reconstruction.

### 3.2.4 *Structural ambiguity*

ACE uses several means to fight structural ambiguity. First, the language does not admit certain ambiguous sentences, or provides unambiguous alternatives. However, not all ambiguous sentences can be eliminated since this would render the language very unnatural. Thus, as a second line of defence structurally ambiguous sentences are parsed deterministically according to a small number of principles associated with syntactic constructions, e.g. *minimal attachment* and *right association*. A paraphrase is generated that shows exactly how the (structurally ambiguous) sentence was parsed. If users find that the paraphrase does not coincide with what they intended to say, they must reformulate the sentence, or decompose the sentence into smaller unambiguous units, e.g. several sentences or a sentence with a relative sentence.

Consider the sentence

```
The customer enters a card with a code.
```

where the prepositional phrase *with a code* could either modify the verb, or the rightmost noun phrase. In ACE only the first alternative – a principle called *minimal attachment* – is realised for the case of prepositional attachments, as reflected by the paraphrase

```
the customer {enters a card with a code}.
```

Since this is probably not what the user intended to express the sentence must be reformulated. To express that the code is attached to the card the user writes

```
The customer enters a card that carries a code.
```

In ACE relative sentences are always attached to the rightmost noun phrase – a principle called *right association*.

Finally, to express that both a card and a code are entered an adequate formulation would be

```
The customer enters a card and a code.
```

Analytical ambiguity arises when the type of a constituent is undecidable. Especially the use of present and past participle forms of verbs leading sometimes to garden-path-sentences are troublesome and therefore forbidden in ACE.

### 3.2.5 *Coordination and Subordination*

In ACE coordination can occur between complete phrasal constituents of equal syntactic status, specifically between sentences. The two most important coordinators are *and* and *or*. Natural language coordination is different from logic coordination where the connectives are commutative. This becomes visible, when we consider the composite sentence

```
The customer enters the card and the customer types a code.
```

The textual order of the two coordinated sentences is decisive because it carries implications about how the events are temporally related. Reversing the order of the sentences leads to another order of events that probably does not correspond to the intentions of the domain specialist.

Another problem turns up when the coordinator *and* joins noun phrases. These coordinated noun phrases enumerate members of a plural set. Plural sentences can have an inherent ambiguity – they can receive a collective or a distributive reading.

```
John and Mary enter a card.
```

Here, it is not clear whether John and Mary enter a card together or separately, leading to one or two entering events, respectively. In ACE one can force a collective reading by using the adverb *together* as keyword.

```
John and Mary enter a card together.
```

A distributive reading is achieved by inserting the adverb *each*.

```
John and Mary each enter a card.
```

In natural language we distinguish two readings of the word *or*, an exclusive and an inclusive one. In ACE we distinguish the two readings by making them explicit; for an exclusive interpretation we write *either ... or*, and for an inclusive interpretation simply *or*.

```
The customer enters either a card or a code or ...
```

```
The customer enters a card or a code or ...
```

To express conditional events and causality the subordinating constructor *if-then* is used.

```
If the code is valid then SimpleMat accepts the card.
```

```
If the code is not valid then SimpleMat rejects the card.
```

Each condition in an *if-then* sentence must be written explicitly, i.e. there is no *else* as in many computer languages.

*3.2.6 Negation*

In ACE we use standard forms to write negative statements so that the same syntactic construction always appears with the same semantic scope of negation.

When a full negative form (*does not*) occurs after the subject to build a verb negation the whole verb phrase falls inside the scope of the negative.

```
The customer does not enter the correct code.
```

When a sentence contains copulative *be*, we form the negative by putting *not* after the copula denying that an identity or a property hold.

```
The card is not valid.
```

The quantifier *no* is the only negative that can occur in a noun phrase. By default it negates an existential and has the whole sentence within its scope.

```
No customer enters a card.
```

When a disjunction is within the scope of a negative the construction is normally interpreted inclusively and denies that all disjuncts are true, e.g.

```
    The customer does not enter a card or a code or ...
```
is interpreted as if the disjunction stands for a conjunction.
```
    The customer does not enter a card and a code and ...
```
To avoid this ambiguous situation, in ACE the user can write for the case that all disjuncts are false
```
    The customer enters neither a card nor a code nor ...
```
or logically equivalent
```
    The customer does not enter a card and does not enter a code and ...
```
```
    The customer does not enter either a card or a code or ...
```
and for the case that some disjuncts are false
```
    The customer does not enter a card or does not enter a code or ...
```

## 4   Overview of Attempto

We have developed the Attempto system that translates ACE specifications into discourse representation structures, and optionally into the logic programming language Prolog. Here, we briefly describe Attempto's components.

The user enters a specification text in ACE that the *Dialog Component* forwards to the parser. Spelling and parsing errors and any remaining ambiguities to be resolved by the user are reported back by the dialog component. The *Parser* uses a predefined grammar and a predefined linguistic lexicon to check sentences for syntactical correctness, and to generate syntax trees and sets of nested discourse representation structures as semantic representation of the input text. The *Linguistic Lexicon* contains a domain-specific vocabulary, and can be modified by a lexical editor invokable from the dialog component.

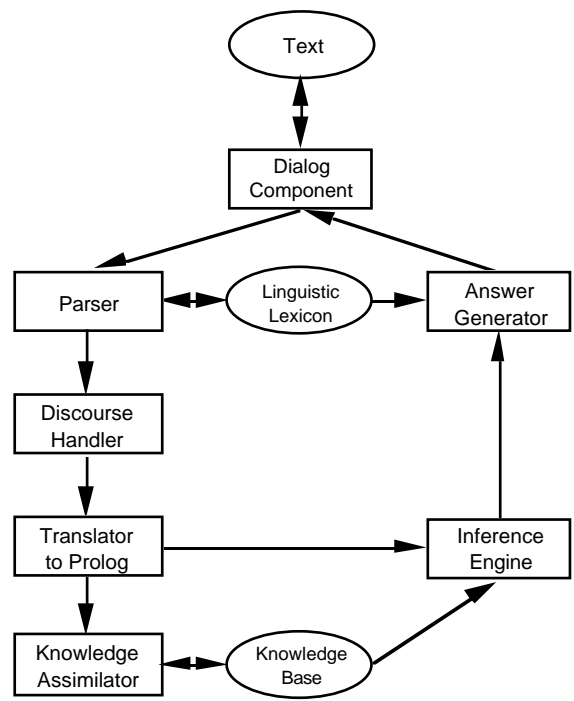

The *Discourse Handler* analyses and resolves inter-text references and updates the discourse representation structures generated by the parser. Optionally, the *Translator* translates discourse representation structures into Prolog clauses. The discourse representation structures and the Prolog clauses are passed to the *Knowledge Assimilator* that adds new knowledge to the *Knowledge Base*.

The user can interrogate the contents of the knowledge base by asking queries in ACE. Translated queries are conveyed to the *Inference Engine* which answers them with the help of the knowledge base. The *Answer Generator* takes the answers of the inference engine, reformulates them in ACE, and forwards them to the dialog component.

# 5 Parsing

The specification text is parsed by a top-down parser using a unification-based phrase structure grammar. The parser builds a syntax tree as syntactic representation, and concurrently a discourse representation structure as semantic representation.

In addition, the parser generates a paraphrase – displaying all substitutions and interpretations made – that explains how Attempto interpreted the user's input. For the sentences

```
The customer enters a card and a numeric personal code.
If it is not valid then SM rejects the card.
```

the paraphrase is

```
the customer enters a card and [the customer enters] a numeric personal code.

if [the personal code] is not valid then [simplemat] rejects the [card].
```

The user can now decide to accept Attempto's interpretation, or to rephrase the input to achieve another interpretation. For ambiguous input Attempto always suggests one standard interpretation as default. It is up to the user to reformulate the input to achieve non-standard interpretations.

Furthermore, the parser informs the user about spelling and parsing errors, and lists unknown words. The user can add unknown words to the lexicon with the help of the lexical editor, and immediately resubmit the input to the parser.

# 6 Semantic Representation

## 6.1 Contextual Semantic Translation

The complete specification text is translated into one discourse representation structure (DRS) which contains discourse referents representing the objects of the discourse, and conditions for these discourse referents [Kamp & Reyle 93].

Each sentence is translated in the context of the preceding sentences, yielding for the sentences

```
The customer enters a card and a numeric personal code.
If it is not valid then SM rejects the card.
```

the DRS

```
[A, B, C, D]
customer(A)
card(B)
enter(A, B)
numeric(C)
personal_code(C)
enter(A, C)
named(D, simplemat)
```

```
      IF:
        []
          NOT:
            []
              valid(C)
        THEN:
          []
            reject(D, B)
```

A DRS is a semantic representation of the input text, and is considered true if the input describes reality. Our example DRS is true if there are objects represented by the discourse referents *A,B,C,D* so that *A* is a customer, *B* is a card, the customer *A* enters the card *B*, etc.

Conditions of a DRS can be simple, e.g. *customer(A)*, or complex, i.e. again a DRS. This can lead to nested DRSs with sub- and superordination. In our case, the topmost DRS contains a subordinate *IF-THEN* DRS which itself contains a subordinate *NOT* DRS.

Anaphoric references, e.g. the pronoun *it* of the second sentence referring to the compound noun *personal code* of the first sentence, are automatically resolved. The resolution algorithm picks the closest referent in a superordinate DRS that agrees in gender and number.

### 6.2 Translation into Prolog

Optionally, the discourse representation structure is translated into Prolog clauses which are added as *fact/1* to the knowledge base.

```
fact(customer(0)).
fact(card(1)).
fact(enter(0, 1)).
fact(numeric(2)).
fact(personal_code(2)).
fact(enter(0, 2)).
fact(named(3, simplemat)).
fact((reject(3, 1):- neg(valid(2)))).
```

In Prolog, the discourse referents *A,B,C,D* – being existentially quantified variables – are replaced by Skolem constants *0, 1, 2, 3*.

## 7 Working with the Knowledge Base

Once we have the specification text translated and the translated form stored in the knowledge base, we can query the knowledge base, or execute it.

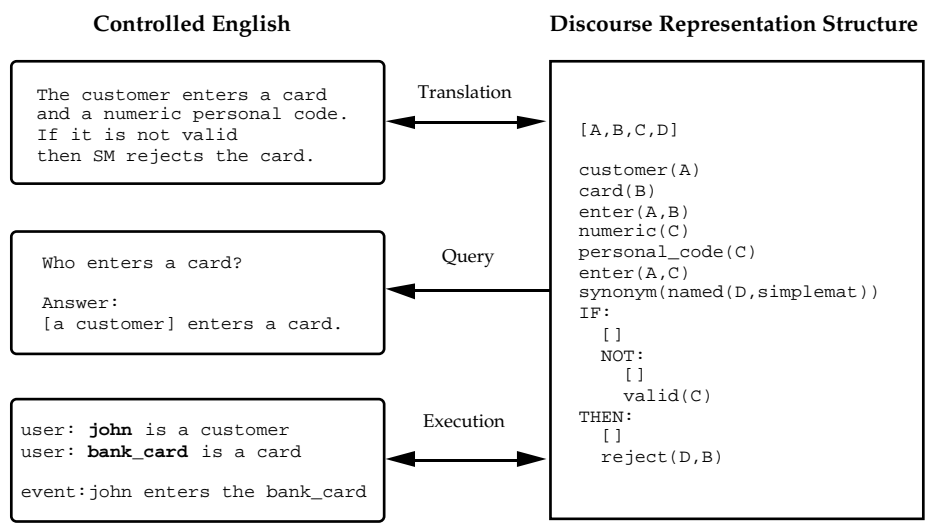

## 7.1 Querying the Knowledge Base

For verification and validation of the specification the knowledge base can be interrogated by queries in ACE. Let us assume that the user entered previously

```
The customer enters a card and a numeric personal code.

If it is not valid then SM rejects the card.
```

Now the user can ask

```
Does the customer enter a card?
```

and the Attempto system will respond with

```
Answer: yes
```

while the query

```
Who enters a card?
```

is answered by the system in a form that shows the substitution

```
Answer: [a customer] enters a card.
```

If a query has more than one answer, all answers can be generated and displayed on demand.

## 7.2 Executing the Specification

The knowledge base can also be used for simulation or prototyping by executing it. In our example specification, this means executing the specification of the automated teller machine. Missing information describing the specific situation and side-effects of events are either defined before-hand in a definition file, or supplied by the user.

Following is an example of the execution of the specification

```
The customer enters a card and a personal code.

SimpleMat checks the personal code.

If the personal code is valid then SimpleMat accepts the card.

If the personal code is not valid then SimpleMat rejects the card.
```

Side-effects of events are simulated by simply printing out a trace of the pertinent event while situation-specific information is provided by querying the user. The execution yields the subsequent dialog where the user's input is written in *italics*

```
user:    john is a customer

user:    bank_card is a card

event:   john enters the bank_card

user:    1234 is a personal_code

event:   john enters 1234

user:    s1 is a simplemat

event:   s1 checks 1234

user:    1234 is not valid

event:   s1 rejects the bank_card
```

# 8   Conclusions and Further Research

The present prototypical implementation of Attempto proves that Attempto Controlled English (ACE) can be used for the non-trivial specification of an automated teller machine, and that the complete specification can be translated as coherent text into a structured form of first-order logic – and optionally into Prolog clauses – that can be executed.

More work needs to be done, however, to complete the definition of ACE, to parse ACE specifications efficiently, and to extend the internal syntactic and semantic representations.